\documentclass[letterpaper]{article} % DO NOT CHANGE THIS
\usepackage{aaai24}  % DO NOT CHANGE THIS
\usepackage{times}  % DO NOT CHANGE THIS
\usepackage{helvet}  % DO NOT CHANGE THIS
\usepackage{courier}  % DO NOT CHANGE THIS
\usepackage[hyphens]{url}  % DO NOT CHANGE THIS
\usepackage{graphicx} % DO NOT CHANGE THIS
\usepackage{natbib}  % DO NOT CHANGE THIS AND DO NOT ADD ANY OPTIONS TO IT
\usepackage{caption} % DO NOT CHANGE THIS AND DO NOT ADD ANY OPTIONS TO IT
\usepackage{algorithm}
\usepackage{babel,blindtext}
\usepackage{algorithmic}
\usepackage{soul}
\usepackage{subcaption}

\usepackage{xcolor}
\usepackage{newfloat}
\usepackage{listings}
\DeclareCaptionStyle{ruled}{labelfont=normalfont,labelsep=colon,strut=off} % DO NOT CHANGE THIS
\floatstyle{ruled}
\newfloat{listing}{tb}{lst}{}
\floatname{listing}{Listing}
\title{LaFFi: Leveraging Hybrid Natural Language \\Feedback for Fine-tuning Language Models}
\author {
    Qianxi Li\textsuperscript{\rm 1}\thanks{Work done during an internship at Huawei Noah's Ark Laboratory, Canada},
    Yingyue Cao\textsuperscript{\rm 1}\footnotemark[1],
    Jikun Kang\textsuperscript{\rm 3},\\
    Tianpei Yang\textsuperscript{\rm 1},
    Xi Chen\textsuperscript{\rm 3},
    Jun Jin\textsuperscript{\rm 1,2},
    Matthew E. Taylor\textsuperscript{\rm 1,2},
}
\affiliations {
    \textsuperscript{\rm 1}University of Alberta, Canada\\
    \textsuperscript{\rm 2}Alberta Machine Intelligence Institute (Amii), Canada\\
    \textsuperscript{\rm 3}Huawei Noah's Ark Laboratory, Canada\\
    \{qianxi, yingyue3, tianpei.yang, jjin5, matthew.e.taylor\}@ualberta.ca, \\\{jaxon.kang,xi.chen4\}@huawei.com}

\usepackage{bibentry}
\begin{document}

\maketitle

\begin{abstract}
Fine-tuning Large Language Models (LLMs) adapts a trained model to specific downstream tasks, significantly improving task-specific performance. Supervised Fine-Tuning (SFT) is a common approach, where an LLM is trained to produce desired answers. However, LLMs trained with SFT sometimes make simple mistakes and result in hallucinations on reasoning tasks such as question-answering. Without external feedback, it is difficult for SFT to learn a good mapping between the question and the desired answer, especially with a small dataset. This paper introduces an alternative to SFT called Natural \textbf{La}nguage \textbf{F}eedback for \textbf{Fi}netuning LLMs (\textbf{\textit{LaFFi}}).  \textbf{\textit{LaFFi}} has LLMs directly \emph{predict} the feedback they will receive from an annotator --- we find that requiring such reflection can significantly improve the accuracy in in-domain question-answering tasks, providing a promising direction for the application of natural language feedback in the realm of SFT LLMs. Additional ablation studies show that the portion of human-annotated data in the annotated datasets affects the fine-tuning performance. 
\end{abstract}

\section{Introduction}

The transformer architecture \cite{attention_is_all_you_needed} has enabled a wide range of applications \cite{vision_transformer, devlin2019bert, raffel2023exploring,sun2019bert4rec}. Large language models (LLMs) have emerged as among the most widely adopted applications founded upon the Transformer architecture due to their remarkable capacity for capturing complex patterns in data and their effectiveness in a diverse array of natural language processing tasks. These LLMs, including models such as GPT \cite{gpt-2,gpt-3}, InstructGPT \cite{instructgpt}, and LLaMA2 \cite{llama2}, have made remarkable advancements in language comprehension tasks \cite{wei2022emergent}. To enhance the accessibility of LLMs, a common practice involves using a model trained on extensive and diverse textual web data and resources as the base model. Subsequently, researchers \cite{wei2022finetuned,Dong2023ASF} only need to fine-tune these pre-trained models on specific downstream tasks to equip them with domain-specific capabilities tailored to user requirements.

Large Language Models (LLMs) trained through Supervised Fine-Tuning (SFT) occasionally exhibit minor errors and generate inaccurate responses in reasoning tasks. In the absence of external guidance, SFT faces challenges in establishing an effective mapping between posed questions and the correct answers, particularly when working with limited datasets. Our approach to fine tune an LLM is to instruct the model to learn the prediction of the natural language feedback it is likely to receive for the responses it generates. There exists an inherent and intuitive connection between human language and the process of training or refining language models, these models are designed to understand and generate human-like text. Natural language feedback brings more information to the training data than the human preference values, which are usually biased \cite{casper2023open}. With an explanation in addition to the desired answer as the label, the LLM learns a better mapping among the input context, the answer it gives, and whether the answer is appropriate and why, thus potentially generating more human-centered and desired responses.

In this work, we present \textbf{\textit{LaFFi}} - Natural \textbf{La}nguage \textbf{F}eedback for \textbf{Fi}netuning Language Models, a framework designed to integrate natural language feedback within the Supervised Fine-Tuning (SFT) paradigm, as shown in Figure \ref{fig:laffi_intro}(a).  This framework is split into four key stages: (1) Answer prediction, (2) Provide natural language feedback to the answers, (3) Use supervised learning to maximize the likelihood of observing the feedback label, (4) Use LoRA \cite{hu2021lora} for parameter-efficient fine-tuning. As far as we know, this is the first work that explores the idea of leveraging natural language feedback by making LLM directly predict what feedback it will receive for certain answers. Our contributions include:
\begin{itemize}
    \item A novel Supervised Fine-Tuning framework --- \textit{LaFFi} --- that fine-tunes an LLM by learning to predict feedback
    \item Using a combination of human and AI annotators for natural language feedback
    \item Experiments to establish that even with a relatively small dataset, \textit{LaFFi} can improve LLM accuracy

\end{itemize}

Our framework \textit{LaFFi}, first uses the LLM to generate answers for the questions, then uses the LLM to directly predict the feedback it receives from the annotators and uses LoRA \cite{hu2021lora} to fine-tune more efficiently, which can significantly improve the accuracy in in-domain question-answering tasks.

\section{Methodology}
\begin{figure}

    \includegraphics[width=0.47\textwidth]{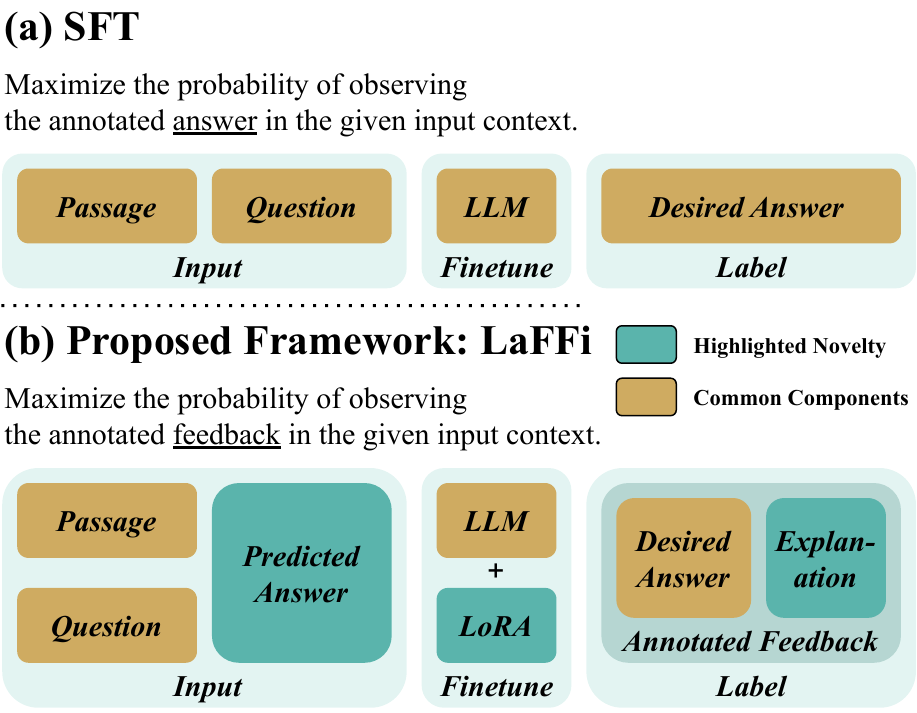}
    \caption{LaFFi is a novel Supervised Fine-Tuning approach that leverages human- and AI-generated free-form natural language feedback along with the parameter-efficient method LoRA to fine-tune LLMs, with low-data requirements. }
    \label{fig:laffi_intro}
\end{figure}

This section introduces a novel LLM fine-tuning framework, \textit{LaFFi}, which contains four steps, as shown in Figure \ref{fig:laffi_intro} (b).
\begin{enumerate}
    \item \textbf{Answer Prediction}. We let our pre-trained LLM generate answers for the questions in the training dataset. We denote the resulting dataset as \textit{Predicted Answer Dataset}.
    \item \textbf{Feedback Annotation}. We use both LLaMA 7B \cite{openlm2023openllama} and human labellers to annotate our \textit{Predicted Answer Dataset}. These annotations include the passage, question, predicted answer, and ground truth answer, with annotators providing feedback in natural language. This feedback is expected to contain details regarding the correct/desired answer, an evaluation of the correctness of our predicted response, and the rationale behind the preference for a particular answer or the shortcomings of our predicted response. We include one example of our training data in the appendix (see Figure~\ref{fig:data_example}). Details regarding the feedback annotation process will be introduced in the Methodology section. The resulting dataset, the \textit{AI/Human Labelled Feedback Dataset}, depends on whether LLM or human labellers annotate it.
    \item \textbf{Supervised Feedback Prediction}. In contrast to SFT, where questions serve as inputs and specific task ground truth answers as labels, \textbf{\textit{LaFFi}} composes a passage, question, and predicted response into a comprehensive input context. The LLM is tasked with predicting the feedback it anticipates for the provided answer. Natural language feedback, collected in advance, is used as the label for this prediction task (see Figure \ref{fig:laffi_intro}~(b)).
    \item \textbf{LoRA Fine-tuning}. We employ the parameter-efficient fine-tuning technique LoRA \cite{hu2021lora} to make training more efficient. This method decomposes specific Multi-Layer Perceptrons (MLPs) \cite{mlp_first_paper} within the Transformer architecture into low-rank matrices, which are then integrated into the MLPs. Simultaneously, other weights in the pre-trained model remain frozen. This approach maximizes the preservation of pre-trained knowledge while reducing computational costs. In our experiments, LoRA is exclusively applied to the \textit{Q\_proj}, \textit{K\_proj}, and \textit{V\_proj} matrices of the Transformer block, affecting only a minimal portion, approximately \textbf{0.09\%}, of the total model parameters that require fine-tuning.
    %, it is an integrated component of our framework.
     
\end{enumerate}

The base dataset used to construct our feedback dataset and in-domain evaluation is \textit{SQuAD 2.0} \cite{squad2.0}, a question-answering dataset that has been widely adopted as a reading comprehension benchmark in NLP. Our datasets are generated from two primary sources: human annotations and AI-generated annotations. We started this process by randomly selecting \textbf{932} instances from the original training dataset. Our choice of AI annotation involves employing the Large Language Model (LLM) LLaMA 7B as the AI labeller, spanning three different model scales: 3B, 7B, and 13B. \textit{LaFFi} first uses a pre-trained LLM to predict the answer for the given prompt that encodes passage and question information, yielding the \textit{Predicted Answer Dataset}. Subsequently, we present the LLM with a question, the answer, the predicted response, and the ground truth answer, instructing it to provide natural language feedback.

For human annotations, the \textit{Predicted Answer Dataset} is randomly divided into six segments, each assigned to a human annotator. To facilitate the annotation process, we supplement each segment with the feedback generated by the LLM. Human annotators are then tasked with providing feedback for the predicted answers. Note that, since not all the answers generated by LLM are accurate, human annotators often need to calibrate their feedback. However, for AI-generated feedback, annotators may simply accept it without further adjustments.

\begin{figure*}[h]

\begin{subfigure}[t]{.242\textwidth}

    \includegraphics[width=\linewidth]{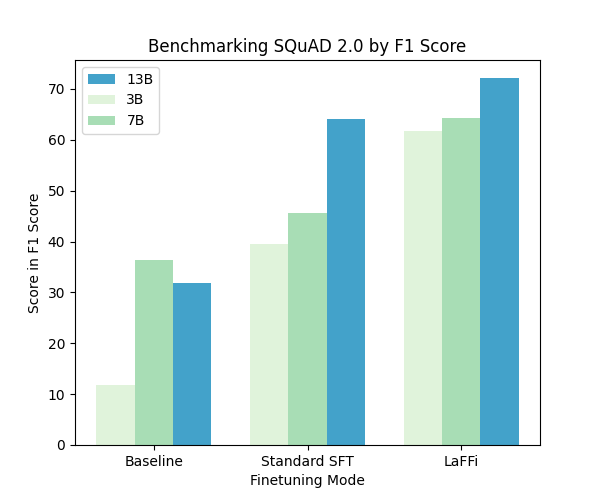}  
    \caption{} %The performance is measured by \textit{F1, Accuracy, Precision} and \textit{Recall}.}
    \label{fig:exp1}
\end{subfigure}
\hspace{0pt}
\begin{subfigure}[t]{.242\textwidth}

    \includegraphics[width=\linewidth]{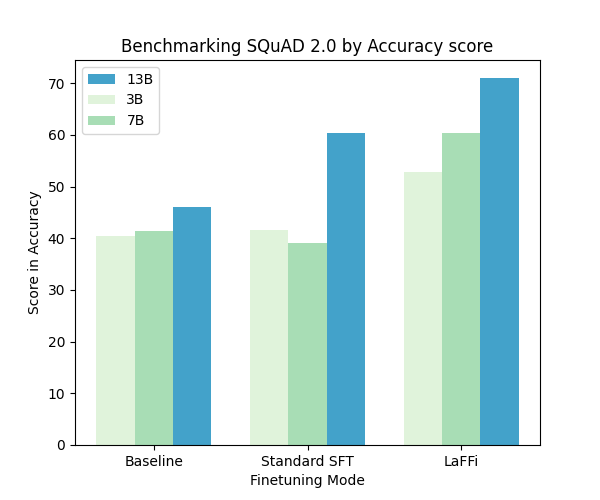}  
    \caption{} %The performance is measured by \textit{F1, Accuracy, Precision} and \textit{Recall}.}

\end{subfigure}
\hspace{0pt}
\begin{subfigure}[t]{.242\textwidth}

    \includegraphics[width=\linewidth]{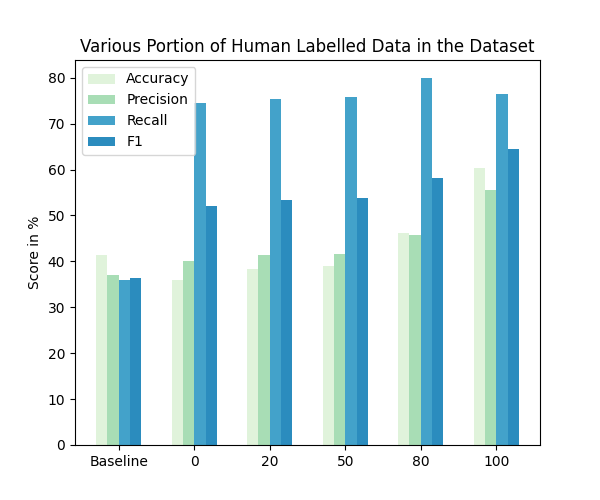}  
    \caption{}
    %The performance is measured in \textit{F1, Accuracy, Precision} and \textit{Recall}.}
    \label{fig:exp2}
\end{subfigure}
\hspace{0pt}
\begin{subfigure}[t]{.242\textwidth}

    \includegraphics[width=\linewidth]{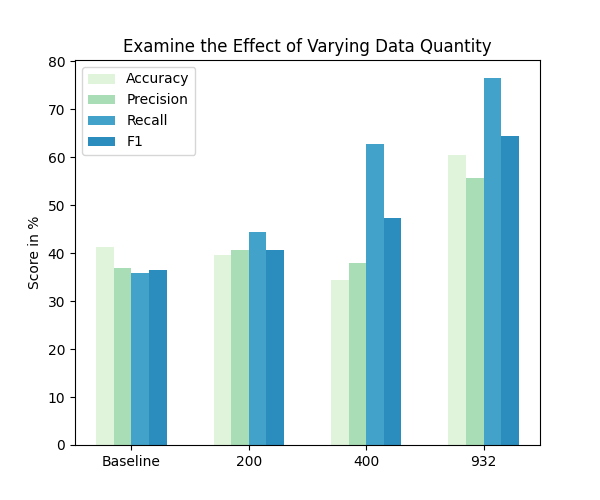}  
    \caption{} %The performance is measured by \textit{F1, Accuracy, Precision} and \textit{Recall}.}
    \label{fig:exp3}
\end{subfigure}
\caption{(a)\&(b) Benchmarking the performance of three model scales (3B, 7B, and 13B) on three different kinds of fine-tuning modes with F1 score and accuracy. (c) X-axis: \textit{Baseline} (no fine-tuning) and \% of human-labeled data from \{0, 20, 50, 80, 100\} (with the remainder of the data coming from AI-labeled data). (d) X-axis: \textit{Baseline} (no fine-tuning) and human labelled feedback (dataset sizes \{200, 400, 932\} rows).}
\label{fig:subfig_all}
\end{figure*}

\section{Experiments}
In this study, we rely on the development set of \textit{SQuAD 2.0} for our evaluations, as access to the official test set is not available to us. All fine-tuned models in the experiments use the two-shot prompt as the input, since \textit{SQuAD 2.0} includes questions for which the answers cannot be derived from the passage, we incorporate examples of both answerable and unanswerable questions.

\subsection{Benchmarking}
We first benchmark the performance of \textit{LaFFi} against the SFT and no fine-tuning, employing three different model scales. The 3B implementation is from OpenLLaMA \cite{openlm2023openllama,together2023redpajama}. The results of our benchmarking process are presented in Figure \ref{fig:exp1}. We fine-tune our LLMs and evaluate once as done in prior work \cite{openai2023gpt4,openlm2023openllama}, so that there is only a single value for each setting (no confidence interval is plotted). \textit{LaFFi} outperforms both the baseline and the SFT approach across all model scales. As the model scale increases, there is a clear performance improvement in terms of both \textit{accuracy} and \textit{F1 Score}. This can be attributed to the larger model's capacity to acquire more complex representations because of its increased parameter count. Additionally, we note a counterintuitive performance decrease in task accuracy for the 3B and 7B scales in the SFT approach. This decline is a consequence of the baseline model's learned tendency to provide ``the answer cannot be found'' for most questions, given the presence of such examples in the prompts. An examination of \textit{precision} and \textit{recall} values reveals their low figures, specifically \textbf{19.23} and \textbf{8.53}, which consequently leads to a low \textit{F1 score}.

\subsection{Human vs.\ AI-Generated Feedback}
Given we have both human-annotated and AI-annotated datasets, a natural question is ``In the context of a relatively modest model scale, what is the relative significance of human-labeled data compared to AI-labeled data, despite the latter's potential sub-optimality, flaws, and errors?'' We generated several datasets with different proportions of human-labeled data while keeping the overall dataset size constant. These subsets of human and AI data were randomly sampled from their respective full datasets, combined, and shuffled. For simplicity, we exclusively employed the annotated dataset from the 7B model. Figure \ref{fig:exp2} offers insights into our observations. Despite the low \textit{recall} and \textit{precision} values, the baseline model predominantly predicts ``the answer cannot be found" resulting in higher \textit{accuracy} compared to fine-tuning with less than 50\% human data. As a larger portion of human data is introduced into the fixed-sized dataset, a noticeable trend of improvement emerges in terms of \textit{F1 score} and \textit{precision}. 
\begin{figure*}[h]
    \centering
    \includegraphics[width=\textwidth]{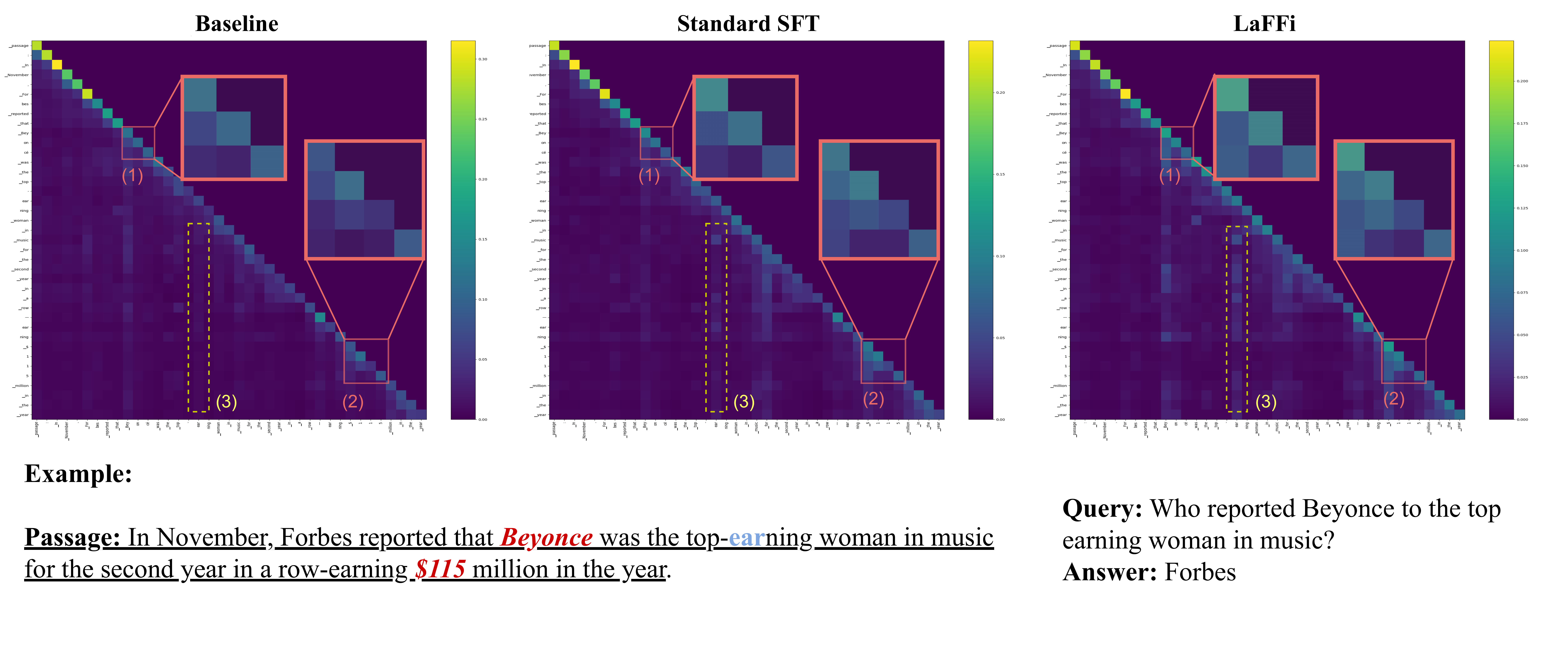}
    \caption{Heatmap visualization of the mean attention score matrix output by the last Transformer block inside LLM. The heatmap for the underlined text is displayed. Warmer colors in the heatmap correspond to higher attention scores between tokens. Regions (1) and (2) on the plots show local dependencies, while (3) shows global dependencies.}
    \label{fig:analysis}
\end{figure*}
\subsection{Quantity}
We also 
%An additional facet of our investigation involved assessing 
assessed how the quantity of data impacted performance.
%necessary to achieve a substantial performance improvement and identify a point of diminishing returns for increasing dataset size.
We considered the 7B human-labeled feedback dataset at three different dataset sizes: 200, 400, and the full 932 items. The results are depicted in Figure \ref{fig:exp3}. The issue that the LLM frequently gives ``the answer cannot be found" discovered in previous experiments still exists in the baseline model and the LLM fine-tuned with 200-rows dataset, but the \textit{F1 score} consistently demonstrates improvement as the dataset size expands. This suggests larger datasets can further enhance performance. 
%enhancements can be achieved by expanding the dataset size beyond 1,000 instances.

\section{Analysis}
%We employ the LoRA method, which involves the integration of low-rank matrices into the \textit{Q\_proj}, \textit{K\_proj}, and \textit{V\_proj} linear projections within each Transformer block during finetuning, while the remaining layers remain frozen.%
To gain deeper insights into the enhancements that have helped our LLM improve performance, a more comprehensive exploration of the underlying Transformer architecture \cite{attention_is_all_you_needed} is necessary. A Transformer block operates by amalgamating information from three matrices and generating attention matrices within each attention head. Inspired by \cite{reid2022wikipedia},  we concentrate on the attention score matrices originating from the last Transformer block within our 7B model. We randomly select a set of $\langle$passage, query, correct answer$\rangle$ triplets from our development dataset and synthesize comprehensive prompts as illustrated in Figure \ref{fig:analysis}. These prompts are subsequently input into the LLM. We compute the average attention score matrix by averaging across 32 attention heads, enabling us to investigate the alterations induced by fine-tuning with LoRA. Upon examination of one of the examples, two noteworthy observations emerge.
\begin{enumerate}
\item \textbf{Strengthened Local Dependencies}. In Figure \ref{fig:analysis}, marked with red bounding boxes, from left to right, we observe a strengthening of local dependencies between highly correlated tokens, the overall brightness increases in these areas in \textit{LaFFi}, compared to baseline and SFT. For instance, the two cases of \textbf{``Beyonce''}, \textbf{``\$115"} in the example are split by the LLM's tokenizer into \textbf{3 and 4} tokens, respectively. However, human comprehension views these as whole terms. Notably, the marked regions exhibit the brightest colors in the \textit{LaFFi} plots, indicating the LLM captured local dependencies.
\item \textbf{Strengthened Global Dependencies}. The yellow, dashed bounding boxes highlight the increased attention towards \textbf{``ear"} in \textbf{``earning"} by other tokens. Comparing three plots, we can see some tokens increase the attention scores on \textbf{``ear''}, such as \textbf{``in music''}, \textbf{``the second year''}, and \textbf{``million''}, as indicated by a brighter color. These tokens are semantically correlated to \textbf{earning}, which is potentially useful for the LLM to capture the semantics of the sentences. The lighter color in these regions signifies a stronger dependency between the token \textbf{``ear"} in \textbf{``earning"} and other non-adjacent tokens in the sentence. This potentially shows that the LLM has developed a more comprehensive understanding of sentence semantics, underscoring its improved grasp of global dependencies.
\end{enumerate}

% \MET{I think we're going to have to work carefully on the readability of these two examples. I'm worried people will quickly get lost and not understand why this is insightful.}
To summarize, we attribute the potential rationale behind the performance enhancements achieved through \textit{LaFFi} to the LLM's refined capabilities in capturing finer token-wise dependencies within the attention blocks.

\section{Related Work}
A few works use human natural language feedback as the feedback source to fine-tune LLMs for different tasks. This section lists a relevant subset of works in this rapidly growing area. REFINER \cite{paul2023refiner} trains a critic model to generate multi-round natural language feedback and iteratively improves the quality of the generator model. \citet{yuan2023systemlevel} provides a framework that leverages natural language feedback to formalize system-level design decisions in a human-in-the-loop process. \citet{openai_self_critique} uses both human and AI annotation and evaluates text summarization tasks, they also consider a similar idea of using critique prediction data as part of the training set, but their experiments require large model scales, massive human data, and hard-to-replicate. PEER \cite{schick2022peer} uses Wikipedia editing and comment history to train a series of encoder-decoder models \cite{encoder_decoder} to build a model to explain why a previous edit was made. 

\section{Conclusion}
We present \textit{LaFFi}, a novel framework that integrates natural language feedback into Supervised Fine-Tuning, while employing parameter-efficient techniques. Extensive experiments on the \textit{SQuAD 2.0} benchmark demonstrate that \textit{LaFFi} delivers substantial performance improvements, surpassing both non-fine-tuned models and Supervised Fine-Tuning, particularly in low-data scenarios. Visualizations highlight \textit{LaFFi}'s ability to capture global and local token dependencies, enhancing few-shot learning.

Limitations include a training dataset that relies exclusively on \textit{SQuAD 2.0}, potentially limiting generalization. Additionally, human annotation is resource-intensive. Future endeavors will address these issues by diversifying datasets, evaluating out-of-domain tasks, and exploring better ways of using AI-generated feedback. We aspire for our work to catalyze further research into the influence of human feedback on Large Language Models within the AI research community.

\section*{Acknowledgments}
Part of this work has taken place in the Intelligent Robot Learning (IRL) Lab at the University of Alberta, which is supported in part by research grants from the Alberta Machine Intelligence Institute (Amii); a Canada CIFAR AI Chair, Amii; Compute Canada; Huawei; Mitacs; and NSERC. 

  \clearpage

\bibliography{aaai24}

\clearpage

\appendix
\section*{Appendix}
\subsection{An example of the training data}
\begin{figure}

    \includegraphics[width=\textwidth]{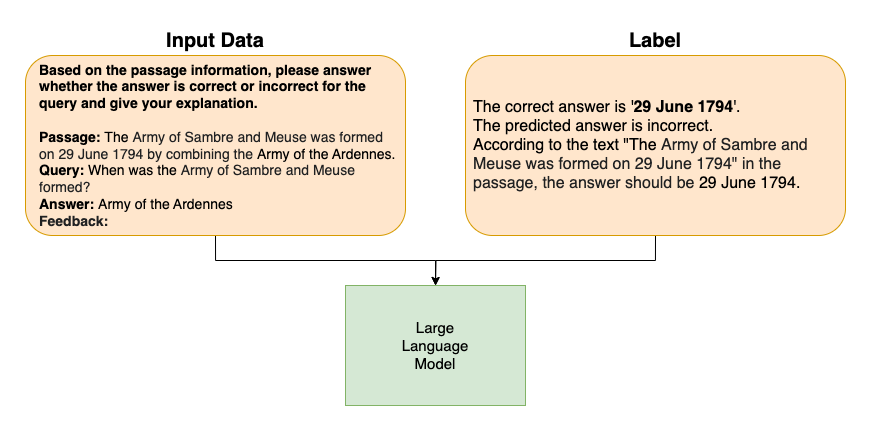}
    \caption{An example of the training data.}
    \label{fig:data_example}
\end{figure}

We provide an example of our training dataset constructed based on the SQuAD 2.0 dataset \cite{squad2.0} in \ref{fig:data_example}. 

The input prompt consists of (1) a necessary prompt context used to make LLM predict feedback. (2) Passage and query question. (3) Predicted answer provided by the LLM to be finetuned. The output label contains the correct answer, the correctness of the predicted answer and the feedback, which includes the rationale for obtaining the correct answer.

\end{document}